# ACCURATE ENTRANCE POSITION DETECTION BASED ON WI-FI AND GPS SIGNALS USING MACHINE LEARNING

**Ahmad Abadleh**
Mutah University; Computer Science Department, JORDAN.
ahmad_a@mutah.edu.jo

***ABSTRACT:*** *This paper aims at detecting an accurate position of the main entrance of the buildings. The proposed approach relies on the fact that the GPS signals drop significantly when the user enters a building. Moreover, as most of the public buildings provide Wi-Fi services, the Wi-Fi received signal strength (RSS) can be utilized in order to detect the entrance of the buildings. The rationale behind this paper is that the GPS signals decrease as the user gets close to the main entrance and the Wi-Fi signal increases as the user approaches the main entrance. Several real experiments have been conducted in order to guarantee the feasibility of the proposed approach. Th experiment results have shown an interesting result and the accuracy of the whole system was one meter.*

**Keywords:** received signal strength (RSS), GPS, machine learning, entrance, signal to noise ratio (SNR).

## I. INTRODUCTION

Indoor localization is one of the main issues nowadays, the importance of indoor localization comes from the diversity of the indoor applications such as healthcare, tracking people inside building, advertisements, and others [1]. Most of the indoor localization techniques need prior knowledge about the buildings such as the blueprint of the building and the Wi-Fi signals. One of the needed information is the position of the entrance which represents a reference for the building. If a user is outside a building while doing tracking and then enters a building, the tracking system must switch from outdoor localization into indoor localization; therefore, the accurate position of the building is essential.

Dynamic indoor localizations detect the position of the users based on landmarks such as Wi-Fi in a specific spot inside the building, unique acceleration in an elevator or stairs [2, 5]. The building entrance can act as a landmark which is used as a reference for localization.

GPS is the main technology used for outdoor localization as the signal is clear. GPS detects the position if at least four satellites are detected with clear signals, and then the trilateration techniques is performed to form multiple spheres in which the intersection area become the position of the user/object [3]. However, GPS signals are very weak and almost unavailable indoors, therefore, the indoor localization techniques utilize Wi-Fi instead of GPS as the Wi-Fi signals available in most of the public buildings [3].

In this paper, we proposed a technique for detecting the entrance based on the GPS signals and the Wi-Fi RSS values. A machine learning algorithm was used with different classifiers in order to detect accurate results. The approach showed the relationship between different features and the entrance of the building. For instance, the number of detected satellites outside the buildings are more than the ones at the entrance. Moreover, GPS signal drops significantly when the user reaches the entrance. If the building offers Wi-Fi then the RSS values increase at the entrance of the buildings.

The main contributions of this paper include the following:
-Studying the relationship between the GPS signals and the entrance of the building.

- Studying the relationship between the Wi-Fi signals and the entrance of the building.
- Studying the relationship between the number of detected satellites and the entrance of the building.
- Utilizing machine learning algorithms to detect accurate entrance position.

## II. RELATED WORK

Many approaches have been proposed in order to solve the problem of indoor localization due to its impotence. The author of [1] proposed an approach to automatically detect the position of Wi-Fi access points. Its approach depends on the strength of the signal and geometry theory to find the closest access point. The approach is important to help in indoor localization as it affects directly the accuracy of the whole system. The proposed approach in [2] solved another important problem in indoor localization which is the distance estimation. Their approach exploits the Hidden Markov Model (HMM) in order to enhance the distance estimation process. An outdoor localization using GPS sensor is enhanced by the approach proposed by [3]. The problem of the outdoor exit is introduced in the paper [4], they use GPS and Wi-Fi signals as well as other sensors to detect the outdoor exit. Their approach utilizes the accelerometer sensor to detect whether the user is moving or not. However; their approach relies on wi-Fi fingerprints which is time-consuming technique.

Smartphone equipped with numerous sensors which can be used to perform many tasks. For instance, the accelerometer sensor can be used to detect the acceleration and the velocity of an object. The authors of [6] proposed an approach to detect the speed bumps on streets, their approach relies on the accelerometer sensor and fuzzy logic inference system. Moreover, the accelerometer sensor used by the approach in [7] to detect the identity of the user. The acceleration is used in forming the encryption key, which is used then in the authentication mechanism.

As there are huge raw data available everywhere, extracting useful information is one of the tasks that machine learning algorithms can do. For instance, [8] provides a comprehensive study about the inference at the edge at Facebook. It can be used in health care [9], society [10] and others.

This paper exploits all the aforementioned technologies to perform detecting the entrance of the buildings. GPS and Wi-Fi sensors which are available in Smartphones are used to collect the data. Several machine learning algorithms applied to the collected data to produce a model which is then used to detect the entrance.

## III. SYSTEM ARCHITECTURE

Figure 1 illustrates the overall system architecture, which consists of the following components: GPS receiver, Wi-Fi receiver, and machine learning algorithms.

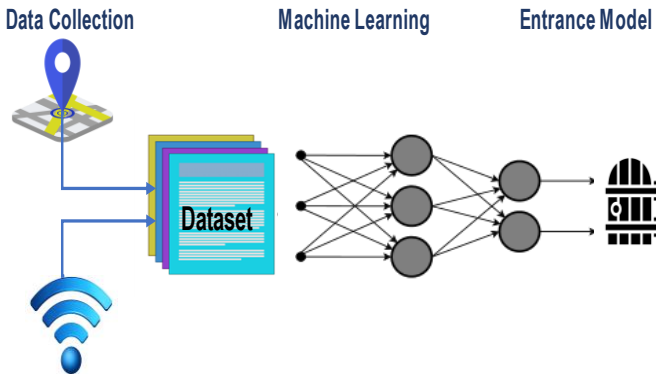

**Figure1 : System Architecture**

As shown in Figure 1, the proposed approach works as follows: the dataset consists of GPS signals and Wi-Fi RSS, then several machine learning algorithms are applied on the dataset to produce a model in which the position of the entrance is detected. The proposed approach consists of three phases as follows: data collection phase, machine learning phase, and entrance model phase.

### A. Data collection phase

In this phase, the readings from GPS sensor and Wi-Fi sensor are collected. The readings include the number of detected satellites, signal to noise ratio (SNR), and Wi-Fi RSS. The readings are then formed the dataset. Table 1 shows a sample of the collected data.

We assume that the states are equiprobable. This assumption is reasonable since it will be the worst scenario compared to trained ones. This means that the state transition probabilities will be equal for all possible states as in Table 1. The important parameter of HMM is the output probability for each state.

Table 1: Sample of collected data

| No of Sat | SNR | RSS | Entrance | Distance | Note |
|---|---|---|---|---|---|
| 20 | 33 | -60 | No | 10 | Outside |
| 14 | 30 | -66 | No | 8 | Outside |
| 23 | 28 | -62 | No | 4 | Outside |
| 15 | 20 | -57 | No | 2 | Outside |
| 9 | 19 | -54 | Yes | 0 | Entrance |
| 8 | 15 | -44 | No | -2 | Inside |
| 4 | 14 | -31 | No | -4 | Inside |

During the collection process, a user holds a Smartphone and record the aforementioned parameters. At each distance, several readings have been stored. For instance, when the user was about 10 m far from the entrance, the Smartphone started to sense the signals from both the GPS and the Wi-Fi. The parameters are selected to be the features for detecting the entrance due to its importance. SNR is used to measure the strength of the signal compared to the noise; therefore; it is a good indicator used to detect whether the signal suffers from high noise or not. A number of detected satellites is a good indicator that there are line-of-sight signals of GPS or not. For instance, if the user inside the building, then there will be no line of sight signal which leads to a smaller number of seen satellites. Wi-Fi RSS used to detect whether a user is closed to an access point or not. For instance, if the signal increases as the user move, this means that the user is going toward an access point.

### B. Machine learning phase

Machine learning as the best solution for extracting the knowledge from raw data becomes an essential topic these days. The proposed approach uses different machine learning classifiers in order to detect the accurate entrance position. The proposed approach applied the following classifiers:

- k-Nearest Neighbor (kNN): kNN is one of the best classifiers as it deals with each case independently, the algorithm measures the distance between the test case and all the existing cases and produce the result based on the k nearest ones.
- Support Vector Machine (SVM): SVM produces the results based on the technique of separating hyperplane.
- Naïve Bayes: Naïve Bayes is a probabilistic classifier in which the results is produced based on the highest probability class.
- Decision Tree: Decision Tree produces the results based on the derived roles from a tree.

### C. Entrance model phase

After the machine learning classifiers are applied on the data, the result of the classifies will form a model. It is used to detect the test cases based on the produced roles. For instance, the decision tree produced the model as roles, such as (if the number of satellites is less than 10 and SNR is low and RSS is high, then the entrance is detected). Figure 2 shows an example of the roles.

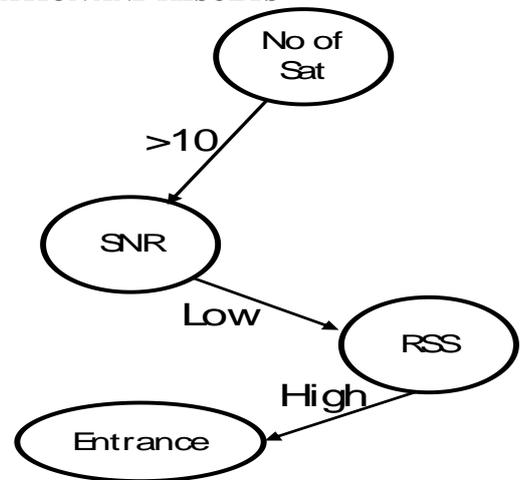

**Figure2 : Decision Tree Sample**

## IV. EVALUATION AND RESULTS

Our proposed approach is tested in a real environment at the department of computer science in Mutah University. In this section, the results of the proposed approach are presented.

### A. Environmental experiments

The experiments were conducted at Mutah University at the Faculty of Science. The user began walking from outside the building until he/she entered the building. The distance from the entrance was 10 m while the user was approaching the entrance. Galaxy S6 Plus Smartphone was used in the experiments and the Wi-Fi analyzer application was used to measure RSS values. GPS test application is used to measure the GPS signals.

### B.    Experiment results

Several real experiments were conducted in order to detect the relationship between some features and the entrance of the building as follows:

Relationship between GPS SNR and entrance position

Figure 3 shows the relationship between SNR and entrance position, the blue shape represents the SNR while the user outside or inside the building and the red shape represents the SNR when the user is at the entrance. As can be seen from the figure, the SNR value is approximately the average between inside and outside which can be an indicator of the entrance position.

.

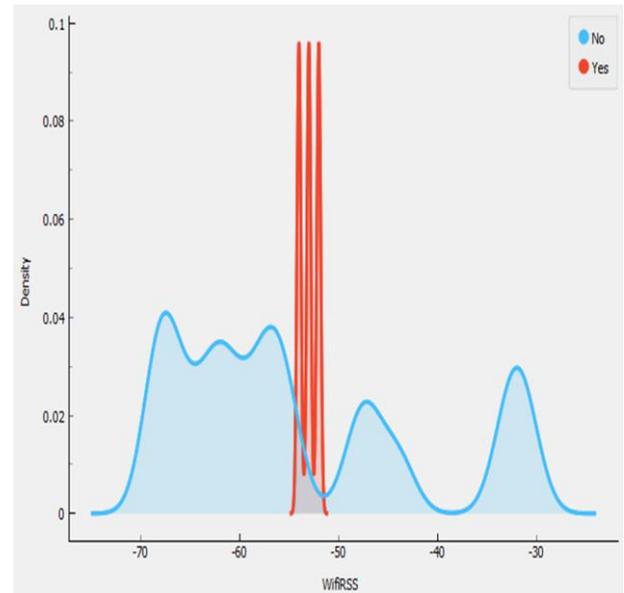

**Figure 4 : Relationship between RSS and entrance position**

**Relationship between RSS and entrance position**

Figure 5 shows the relationship between the number of detected satellites and the entrance position, the blue shape represents the number of satellites while the user outside or inside the building and the red shape represents the number when the user is at the entrance. As can be seen from the figure, the represents the number value is approximately the average between inside and outside which can be an indicator of the entrance position.

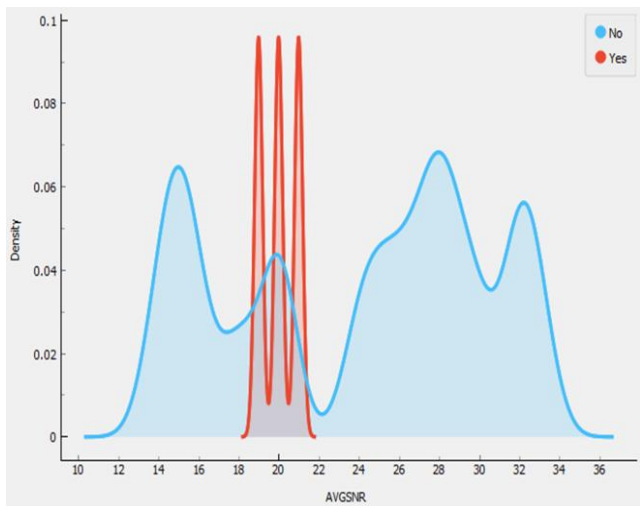

**Figure3 : Relationship between GPS SNR and entrance position**

**Relationship between RSS and entrance position**

Figure 4 shows the relationship between RSS and entrance position, the blue shape represents the RSS while the user outside or inside the building and the red shape represents the RSS when the user is at the entrance. As can be seen from the figure, the RSS value is approximately the average between inside and outside which can be an indicator of the entrance position.

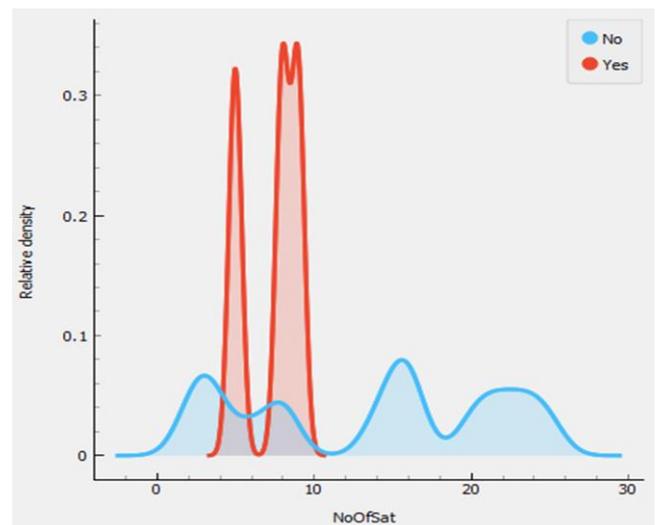

**Figure5 : Relationship between #of Satellites and entrance position**

**Proposed approach accuracy**

Figure 6 shows the results of the classifiers where four classifiers were used. All the classifiers achieved the results on less than one meter. As seen in the figure, kNN achieved 95% accuracy which is the highest and Naïve Bayes achieved 81% which the lowest. These results proved the feasibility of the proposed approach in detecting the entrance of the building.

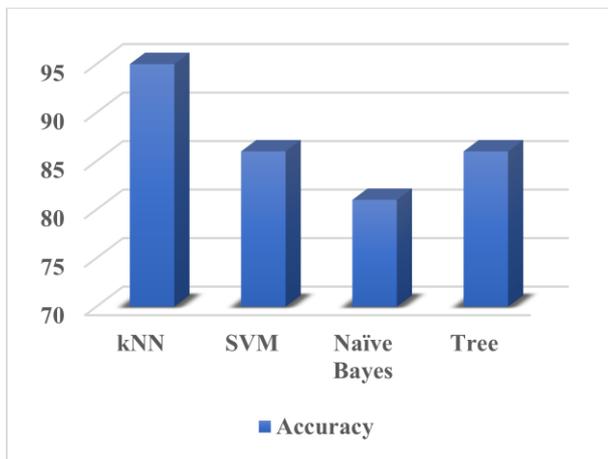

**Figure 6:** Classifiers' Accuracy

## V. CONCLUSION

Detecting the entrance of the building helps in many issues such as indoor localization. In this paper, we proposed a technique for detecting the entrance based on the GPS signals and the Wi-Fi RSS values. A machine learning algorithm was used with different classifiers in order to detect accurate results. The approach showed the relationship between different features and the entrance of the building. For instance, the number of detected satellites outside the buildings are more than the ones at the entrance. Moreover, GPS signal drops significantly when the user reaches the entrance. If the building offers Wi-Fi then the RSS values increase at the entrance of the buildings. Experiment results showed that the approach achieved high accuracy within one meter.

## REFERENCES

[1] Ahmad Abadleh. Wi-Fi RSS-Based Approach for Locating the Position of Indoor Wi-Fi Access Point. Communications - Scientific Letters of the University of Zilina, 21(4), 2019, 69-74.

[2] Ahmad Alabadleh, Saqer Aljaafreh, Ahmad Aljaafreh & Khaled Alawasa (2018) A RSS-based localization method using HMM-based error correction, Journal of Location-Based Services, 12:3-4, 2018, 273-285.

[3] Santerre, R., Geiger, A., Banville, S., Geometry of GPS relative positioning. GPS Solut, 2018.

[4] L.M. Soria Morillo, J.A. Ortega Ramírez, J.A. Alvarez García, L. Gonzalez-Abril, Outdoor exit detection using combined techniques to increase GPS efficiency, Expert Systems with Applications,Volume 39, Issue 15, 2012, Pages 12260-12267.

[5] A. Murugan, R. and R. , "Localization based User Tracking Using RSSI," in International Conference on Innovations in Engineering and Technology (ICIET'16), India, 2016.

[6] Ahmad Aljaafreh, Khaled Alawasa, Saqer Alja'afreh, Ahmad Abadleh, Fuzzy Inference System for Speed Bumps Detection Using Smart Phone Accelerometer Sensor, Journal of Telecommunication, Electronic and Computer Engineering, Vol 9, No 2-7, 2017.

[7] S. Mulhem, A. Abadleh and W. Adi, "Accelerometer-Based Joint User-Device Clone-Resistant Identity," 2018 Second World Conference on Smart Trends in Systems, Security and Sustainability (WorldS4), London, 2018, pp. 230-237.

[8] C. Wu et al., "Machine Learning at Facebook: Understanding Inference at the Edge," 2019 IEEE International Symposium on High Performance Computer Architecture (HPCA), Washington, DC, USA, 2019, pp. 331-344.

[9] Beam AL, Kohane IS. Big Data and Machine Learning in Health Care. JAMA. 2018;319(13):1317–1318.

[10] Eric Xing. 2018. SysML: On System and Algorithm Co-design for Practical Machine Learning. In Proceedings of the 24th ACM SIGKDD International Conference on Knowledge Discovery & Data Mining (KDD '18). ACM, New York, NY, USA, 2880-2880, 2018.